\documentclass[runningheads]{llncs}

\usepackage{amsmath}
\usepackage{booktabs}
\usepackage{amssymb}
\usepackage[table]{xcolor}
\usepackage[T1]{fontenc}
\usepackage{enumitem}
\usepackage{amssymb}
\usepackage{threeparttable}
\usepackage{multirow}
\usepackage{graphicx} 
\usepackage{xcolor}
\usepackage[most]{tcolorbox}
\usepackage{tabularx}
\usepackage{pifont}
\definecolor{mybordercolor}{HTML}{545454}
\definecolor{mybgcolor}{HTML}{F5F5F5}
\tcbuselibrary{skins} 

\usepackage{graphicx,verbatim}
\begin{document}
\title{LogicEye: Agent Driven Multimodal Dataset for Interpretable Ophthalmic Diagnosis}
%

\author{
Jiajie Mu\inst{1} \and
Lu Chen\inst{1} \and
Zheyun Qin\inst{2} \and
Xiaoming Xi\inst{1}\thanks{Corresponding author.} \and
Lishan Qiao\inst{1} \and
Xiushan Nie\inst{1} \and
Yilong Yin\inst{2}
}

\authorrunning{J. Mu et al.}

\institute{
School of Computer and Artificial Intelligence,
Shandong Jianzhu University, Jinan 250101, China\\
\email{edensmjjz@gmail.com, astrojackson825@gmail.com}\\
\email{fyzq10@126.com, qlishan@163.com, niexiushan@163.com}
\and
School of Software,
Shandong University, Jinan 250101, China\\
\email{zheyunqin@gmail.com, ylyin@sdu.edu.cn}
}
  
\maketitle              

\begin{abstract}
Clinical adoption of ophthalmic Artificial Intelligence remains limited by low interpretability and a lack of finely annotated multimodal data. Most existing datasets provide only coarse disease labels without the granular pathological findings required for differential diagnosis. We present LogicEye$^{Agent}$, a dual-pool inference and verification agent that emulates clinical reasoning by jointly verifying positive inclusion signs and negative exclusion criteria. Using LogicEye$^{Agent}$, we curate LogicEye$^{Data}$, a large-scale multimodal dataset of 804,461 fundus and OCT images paired with structured pathological reports spanning 38 diseases and 29 subtypes, with quality assured through joint verification by multimodal large language models and clinical experts. Building on this dataset, we further develop LogicEye$^{Bench}$, which incorporates a category-constrained loss for hierarchical disease recognition and report generation. Grounding coarse labels into verified visual findings effectively curtails diagnostic hallucinations and establishes a solid foundation for interpretable vision-language modeling in ophthalmology. The code is available at:https://github.com/Eden35/LogicEye-Agent-Driven-Multimodal-Dataset-for-Interpretable-Ophthalmic-Diagnosis
\end{abstract}
\section{Introduction}

Early screening and accurate diagnosis are essential for preventing irreversible vision loss from ophthalmic diseases. Artificial intelligence (AI) has achieved expert-level performance in screening for diabetic retinopathy, glaucoma, and age-related macular degeneration~\cite{zhou2023foundation,de2023airogs}. Multimodal large language models (MLLMs)~\cite{li2023llava,zhang2023biomedclip,thirunavukarasu2023large} have extended ophthalmic AI beyond disease classification to broader image--language understanding~\cite{silva2025foundation,deng2024ophglm,du2024ret}. Yet high classification accuracy does not ensure that a model can explain which image findings support its diagnosis. Interpretability, trust, and generalization remain challenges for clinical use~\cite{wu2025eyecare,zhu2025can,kim2024transparent}.

The available training data provide limited support for this level of interpretation. Many public ophthalmic datasets contain thousands to tens of thousands of images and cover a single modality, such as fundus photography or optical coherence tomography (OCT)~\cite{khan2021global,wang2024mcpl,liu2026leveraging}. Their annotations also tend to provide coarse disease labels~\cite{jin2022fives,tian2025coarse,mok2022affine,jiang2026pathreasoner}, without structured descriptions of disease subtypes and pathological signs~\cite{tian2025coarse,mok2022affine}. Disease labels alone provide little guidance on the visual findings that support a diagnosis or exclude alternatives. Without this supervision, generated reports can contain unsupported findings~\cite{qu2025pncd,asgari2025framework}, reducing their reliability for clinical decision support~\cite{zhu2025can,lim2023benchmarking}.

\begin{figure}[t]
  \centering
  \includegraphics[width=\textwidth]{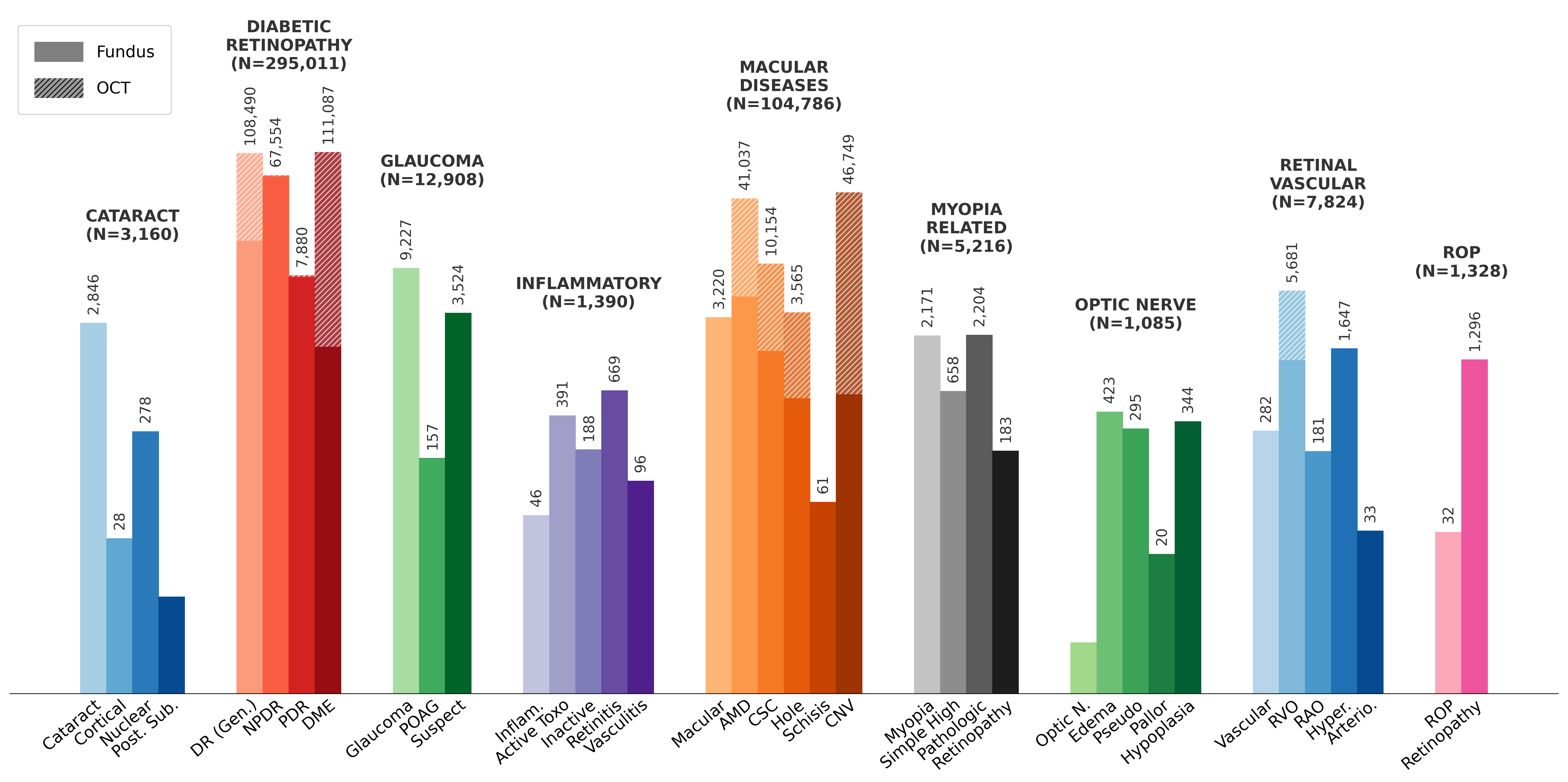} 
  \vspace{-20pt}
  \caption{Distribution of LogicEye$^{Data}$. Numbers above each category indicate the sample count ($N$).}
  \label{fig:data_distribution}
  \vspace{-8pt}
\end{figure}

We introduce LogicEye$^{Data}$, a multimodal ophthalmic dataset that pairs fundus and OCT images with structured pathological diagnostic reports. It brings together 804,461 images from 67 public sources: 346,649 color fundus photographs and 457,812 OCT images (Table~\ref{tab:raw_data_full}). The dataset covers 9 major disease categories and 29 fine-grained subtypes (Fig.~\ref{fig:data_distribution}).

To construct these reports, LogicEye$^{Agent}$ retrieves disease-specific criteria from a structured expert knowledge base covering 38 ophthalmic diseases. A positive feature pool supplies inclusion criteria, while a negative feature pool supplies exclusion criteria. The agent checks candidate features against each image and records verified findings and relevant absent signs in the report. This links disease labels to explicit visual evidence for differential diagnosis. Annotation quality is assessed through automated MLLM evaluation and clinical expert review. LogicEye$^{Bench}$, based on Qwen3 VL~\cite{yang2025qwen3technicalreport}, uses a category-constrained loss to combine report generation with classification of major disease categories and fine-grained subtypes.

Our contributions are as follows:
\vspace{-8pt}
\begin{enumerate}
\item LogicEye$^{Data}$ contains 804,461 fundus and OCT images paired with structured pathological diagnostic reports, providing fine-grained supervision for ophthalmic vision-language modeling.
\item LogicEye$^{Agent}$ uses dual-pool retrieval and visual verification to annotate positive inclusion signs and negative exclusion criteria for differential diagnosis.
\item An annotation verification framework combines automated MLLM evaluation with clinical expert review to assess the accuracy and consistency of generated reports.
\item LogicEye$^{Bench}$ combines report generation with a category-constrained loss for joint recognition of major disease categories and fine-grained subtypes.
\end{enumerate}

\begin{table}[t]
\centering
\caption{Source datasets and image counts.}
\label{tab:raw_data_full}
\vspace{-4pt}
\scriptsize 
\renewcommand{\arraystretch}{1.1} 
\setlength{\tabcolsep}{1.5pt} 

\begin{tabular}{llr @{\hspace{6pt}} | @{\hspace{6pt}} llr @{\hspace{6pt}} | @{\hspace{6pt}} llr}
\toprule
\textbf{Mod} & \textbf{Dataset} & \textbf{Count} & 
\textbf{Mod} & \textbf{Dataset} & \textbf{Count} & 
\textbf{Mod} & \textbf{Dataset} & \textbf{Count} \\
\midrule
Fun & DRA / Eyepacs & 70,252 & Fun & ROP Infant & 1,286 & Fun & DR1-2 & 130 \\
Fun & Diabetic Re & 35,093 & Fun & SUSTech-SYSU & 1,219 & Fun & DRIONS-DB & 110 \\
Fun & DR1-5 & 34,043 & Fun & PALM & 1,200 & Fun & DRISHTI-GS & 101 \\
Fun & JICHI & 19,878 & Fun & SUSTech & 1,151 & Fun & HRF & 90 \\
Fun & DeepEyeNet & 19,407 & Fun & G1020 & 1,020 & Fun & DiaRetDB1 & 89 \\
Fun & AIROGS & 18,000 & Fun & Harvard G & 1,000 & Fun & RITE & 80 \\
Fun & AOD & 14,813 & Fun & JSIEC & 925 & Fun & DR-Hagis & 40 \\
Fun & iChallenge-2018 & 14,145 & Fun & FIVES & 800 & Fun & DRHAGIS & 40 \\
Fun & OIA-DDR & 12,522 & Fun & PARAGUAY & 757 & Fun & DRIVE & 40 \\
Fun & BRSET & 8,775 & Fun & ACRIMA & 705 & Fun & AGAR300 & 28 \\
Fun & Glaucoma & 7,815 & Fun & ORIGA & 650 & Fun & CHASEDB1 & 28 \\
Fun & ODIR & 7,439 & Fun & BEH & 634 & Fun & LES-AV & 22 \\
Fun & APTOS & 5,590 & Fun & Retina & 601 & \multicolumn{3}{c}{\cellcolor{gray!10}\textit{-- Fundus Subtotal --}} \\
Fun & LAG & 4,854 & Fun & DRIMDB & 584 & \multicolumn{3}{c}{\textbf{346,649}} \\
Fun & IDRiD & 4,256 & Fun & PAPILA & 488 & OCT & OCTA-500 & 180,800 \\
Fun & RETOUCH & 3,939 & Fun & RIM-ONE DL & 485 & OCT & OLIVES & 161,047 \\
Fun & BiDR & 2,838 & Fun & E-ophta & 442 & OCT & OCT2017 & 83,599 \\
Fun & MuReD & 2,638 & Fun & ToxoFundus & 411 & OCT & Retinal OCT-C8 & 23,997 \\
Fun & Messidor-2 & 2,471 & Fun & RFMiD-2 & 410 & OCT & APTOS (OCT) & 5,590 \\
Fun & DeepDRiD & 1,589 & Fun & iChallenge-AMD & 400 & OCT & AMD-SD & 2,779 \\
Fun & MMAC & 1,572 & Fun & ODIR-5K & 313 & \multicolumn{3}{c}{\cellcolor{gray!10}\textit{-- OCT Subtotal --}} \\
Fun & Kim's Eye & 1,369 & Fun & Retinal Occlusion & 278 & \multicolumn{3}{c}{\textbf{457,812}} \\
Fun & CHAKSU & 1,345 & Fun & FUND / ARIA & 323 & \multicolumn{3}{c}{\cellcolor{gray!20}\textbf{TOTAL: 804,461}} \\
\bottomrule
\end{tabular}
\vspace{-10pt}
\end{table}

\section{Related Work}

Ophthalmic data resources have expanded from task-specific annotations to datasets for pretraining and image--text learning. FIVES~\cite{jin2022fives} provides pixel-level vessel annotations for retinal image segmentation. RetinaLogos~\cite{ning2025retinalogos} provides a captioned fundus dataset with 1.4 million entries for text-driven image synthesis. RETFound~\cite{zhou2023foundation} uses masked autoencoders (MAE) to learn visual representations from unlabeled fundus and OCT images. FLAIR~\cite{silva2025foundation} incorporates expert knowledge of pathological features through text supervision. RetiZero~\cite{wang2025enhancing} combines MAE and CLIP with knowledge of more than 400 fundus diseases for zero-shot recognition.

These resources provide supervision for segmentation, image synthesis, and representation learning. LogicEye$^{Data}$ adds structured pathological reports that link disease labels to verified visual findings and relevant absent signs for differential diagnosis. Its hierarchical labels support recognition of major disease categories and fine-grained subtypes. Table~\ref{tab:comparison} compares the scale, modalities, annotations, and label organization of these resources.

\begin{table}[t]
\centering
\caption{Comparison of ophthalmic datasets and pretraining resources. F: fundus; O: OCT. L: disease labels; R: diagnostic reports.}
\label{tab:comparison}
\vspace{-8pt}
\begin{threeparttable}
    \begin{tabular*}{\textwidth}{l @{\extracolsep{\fill}} c c c c c}
    \toprule
    \textbf{Dataset} & \textbf{Modality} & \textbf{Scale} & \textbf{Annotation} & \textbf{Hierarchy}\tnote{$\dagger$} & \textbf{Classes}  \\
    \midrule
    FIVES         & F       & 0.8k  & L                 & No         & 3         \\
    RetinaLogos   & F       & 1.4M  & L \& R          & No         & -      \\
    RETFound      & F \& O     & 1.6M  & L                  & No         & 39        \\
    FLAIR         & F       & 288k  & L \& R          & No         & 101\tnote{*}  \\
    RetiZero      & F       & 341k  & L                 & No         & 400\tnote{*}  \\
    \midrule
    \textbf{LogicEye$^{Data}$} & F \& O     & 804k & L \& R & Yes & 38/304\tnote{*}  \\
    \bottomrule
    \end{tabular*}

    \begin{tablenotes}
        \footnotesize
        \item[*] Counts may combine diseases and fine-grained features. LogicEye$^{Data}$ includes 38 diseases and 304 fine-grained features.
        \item[$\dagger$] Hierarchy indicates whether major disease categories and fine-grained subtypes are represented separately.
    \end{tablenotes}
\end{threeparttable}
\vspace{-8pt}
\end{table}
\section{LogicEye$^{Data}$ and LogicEye$^{Agent}$}
\label{sec:dataset}

\begin{figure}[htb!]
\centering
\includegraphics[width=\textwidth]{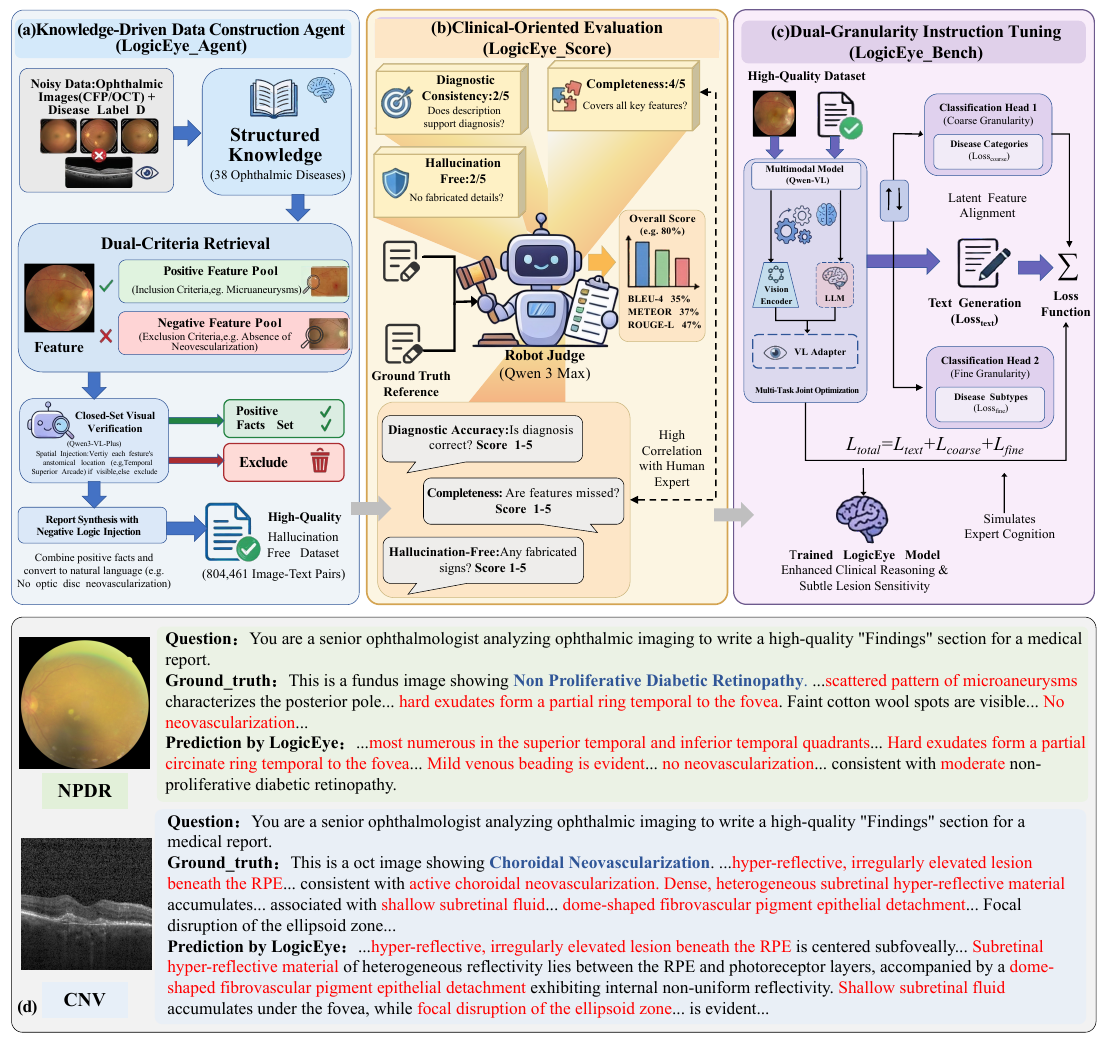}
\vspace{-20pt}
\caption{LogicEye framework. (a) LogicEye$^{Agent}$ retrieves inclusion and exclusion criteria and checks candidate features against each image to generate a structured report. (b) LogicEye$^{Score}$ uses an MLLM to assess diagnostic accuracy, feature completeness, and non-hallucination. (c) LogicEye$^{Bench}$ combines report generation with coarse- and fine-grained disease classification. (d) Example predictions for fundus and OCT images.}
\label{fig:framework}
\vspace{-10pt}
\end{figure}

LogicEye$^{Data}$ contains 804,461 images: 346,649 color fundus photographs and 457,812 optical coherence tomography (OCT) images. The dataset includes normal images, 9 major disease categories, and 29 fine-grained subtypes (Fig.~\ref{fig:data_distribution}). LogicEye$^{Agent}$ (Dual-pool inference \& Verification Agent) generates structured pathological reports by checking disease-specific visual criteria against each image.

LogicEye$^{Agent}$ uses a structured expert knowledge base $\mathcal{K}$ covering 38 ophthalmic diseases (Fig.~\ref{fig:pipeline}). For a given disease label $D$, it retrieves a positive feature pool $\mathcal{P}_{D}$ and a negative feature pool $\mathcal{N}_{D}$. The positive pool contains inclusion criteria, such as microaneurysms and hard exudates. The negative pool defines exclusion criteria for differential diagnosis and staging, such as the absence of neovascularization in nonproliferative diabetic retinopathy.

\begin{figure}[htb]
\centering
\begin{tcolorbox}[
    colframe=mybordercolor,
    colback=mybgcolor,
    arc=5mm,
    boxrule=1pt,
    width=\textwidth,
    top=2pt, bottom=2pt, left=4pt, right=4pt ]
\scriptsize
\textbf{1. Structured Knowledge Base ($\mathcal{K}$) [Source: NPDR Fundus JSON]} \par
\vspace{2pt}

\begin{itemize}[label=\scriptsize$\blacktriangleright$, leftmargin=*, itemsep=1pt, parsep=0pt]
    \item "Scattered \textbf{microaneurysms} visible in the posterior retina..."
    \item "Multiple \textbf{dot-and-blot hemorrhages} within the retina..."
    \item "\textbf{Hard exudates} (waxy yellow white with distinct margins)..."
    \item "Retinal veins may show focal dilation or beading..."
\end{itemize}

\vspace{3pt}
\textbf{2. Derived Dual-Feature Criteria} \par
\vspace{2pt}
\textbf{Positive Feature Criteria ($\mathcal{P}$) [Inclusion]:} Microaneurysms; Dot-and-blot hemorrhages; Hard exudates; Cotton wool spots; Venous beading. \par
\vspace{2pt}
\textbf{Negative Feature Criteria ($\mathcal{N}$) [Exclusion]:} \textit{No} Neovascularization (NVD/NVE); \textit{No} Vitreous Hemorrhage; \textit{No} Tractional Detachment.
\end{tcolorbox}
\vspace{-9pt}
\caption{Example knowledge-base entry and derived inclusion and exclusion criteria for nonproliferative diabetic retinopathy (NPDR).}
\label{fig:pipeline}
\vspace{-15pt}
\end{figure}

LogicEye$^{Agent}$ uses a vision-language model to check each candidate feature against the image. For visible features, it records the anatomical location and a visual description. Features judged absent from the image are recorded as negative findings. For a normal fundus image, the report may state ``no hemorrhage observed,'' ``no exudate observed,'' and ``clear optic disc margins.'' The resulting report links the disease label to image-specific pathological findings and descriptions of absent signs.

An MLLM checks each image--report pair for diagnostic consistency and assigns a score from 1 to 5 on each of three dimensions:
\begin{enumerate}
    \item Diagnostic Accuracy ($S_{acc}$): Assesses whether the report correctly identifies the major disease category and fine-grained subtype.
    \item Feature Completeness ($S_{comp}$): Assesses whether the report includes key pathological findings and their attributes, such as hemorrhage location or fluid morphology.
    \item Non-hallucination ($S_{hall}$): Assesses whether reported findings are visible in the image, with higher scores indicating fewer unsupported findings.
\end{enumerate}

LogicEye$^{Score}$ combines the three component scores as a weighted average:
\begin{equation}\small
    \text{LogicEye$^{Score}$} = \gamma_1 S_{acc} + \gamma_2 S_{comp} + \gamma_3 S_{hall}
\end{equation}
where $\gamma_i \geq 0$ and $\sum_{i=1}^3 \gamma_i = 1$, keeping the aggregate score on a 1--5 scale. Samples with $\text{LogicEye$^{Score}$}<2$ are flagged for manual review. For disputed cases, the initial annotation and reviewed version are submitted to senior ophthalmologists, who determine the final annotation.

\vspace{-8pt}
\section{LogicEye$^{Bench}$}
\vspace{-4pt}

LogicEye$^{Bench}$ uses dual-granularity instruction tuning to jointly learn diagnostic report generation and disease classification. Two classification tasks provide supervision for 9 major disease categories and 29 fine-grained subtypes, respectively. This objective encourages the model to retain disease-discriminative features while generating reports.

We use the MLLM's final-layer hidden state, denoted by $H_{multi}$, as the shared input to two lightweight classification projection heads. One head predicts the major disease category, and the other predicts the fine-grained subtype. We jointly optimize report generation and both classification tasks with the following loss:
\begin{equation}\small
    \mathcal{L}_{total} = \mathcal{L}_{gen} + \lambda_1 \mathcal{L}_{coarse} + \lambda_2 \mathcal{L}_{fine}
\end{equation}
where $\mathcal{L}_{gen}$ is the standard autoregressive cross-entropy loss for report generation. The weights $\lambda_1$ and $\lambda_2$ control the contributions of the coarse- and fine-grained classification losses. Both classification tasks use cross-entropy, defined for a task with $K$ classes as:
\begin{equation}\small
    \mathcal{L}_{task} = -\sum_{k=1}^{K} \mathbb{I}(y_{gt} = k) \cdot \log P(k | H_{multi})
\end{equation}
where $y_{gt}$ is the ground truth label and $\mathbb{I}(\cdot)$ is the indicator function. We set $K=9$ for coarse-grained classification and $K=29$ for fine-grained classification. These auxiliary losses provide direct supervision for disease recognition at both levels during report generation.

\textbf{Implementation Details.} We initialize LogicEye$^{Bench}$ from Qwen3 VL 8B for instruction tuning. During the initial training stages, the visual encoder is frozen, while the cross-modal adapter and LLM parameters are fine-tuned. We use AdamW with an initial learning rate of $1 \times 10^{-5}$ and a cosine annealing schedule. The model is trained for 10 epochs on 8 NVIDIA A100 GPUs with a batch size of 4. The dataset is split into training, validation, and test sets in an 8:1:1 ratio.

\vspace{-8pt}
\section{Experiments}
\label{sec:experiments}
\vspace{-8pt}

\noindent \textbf{Performance Comparison.}
We compared LogicEye$^{Bench}$ with three commercial MLLMs on the LogicEye$^{Data}$ test set (Table~\ref{tab:model_compare}). LogicEye$^{Bench}$ achieved the highest values across all four metrics, including a LogicEye$^{Score}$ of 2.92 and an ACC of 0.54. GPT 4o, the strongest commercial baseline on these two metrics, scored 2.11 and 0.32, respectively. Category-level ACC was 0.83 for optic nerve disorders and 0.71 for diabetic retinopathy (Table~\ref{tab:per_class_acc}).

\begin{table}[t]
\centering
\caption{Comparison of LogicEye$^{Bench}$ and three commercial MLLMs on the LogicEye$^{Data}$ test set.}
\vspace{-4pt}
\label{tab:model_compare}
\scriptsize
\setlength{\tabcolsep}{3pt}
\begin{tabular}{lccccc}
\toprule
\textbf{Model} & \textbf{Params} & \textbf{LogicEye$^{Score}$} & \textbf{ACC} & \textbf{METEOR} & \textbf{ROUGE-L} \\
\midrule
LogicEye$^{Bench}$ & 8B & \textbf{2.92} & \textbf{0.54} & \textbf{0.47} & \textbf{0.44} \\
GPT 4o~\cite{openai2024gpt4ocard} & Proprietary & 2.11 & 0.32 & 0.28 & 0.27 \\
Gemini 3.0 Flash~\cite{comanici2025gemini25pushingfrontier} & Proprietary & 2.05 & 0.28 & 0.26 & 0.25 \\
Qwen3 VL Plus~\cite{yang2025qwen3technicalreport} & Proprietary & 1.80 & 0.25 & 0.24 & 0.23 \\
\bottomrule
\end{tabular}
\vspace{-4pt}
\end{table}

\begin{table}[t]
\centering
\caption{Classification accuracy across the nine major disease categories.}
\label{tab:per_class_acc}
\vspace{-4pt}
\scriptsize
\setlength{\tabcolsep}{4pt}
\begin{tabular}{lccccccccc}
\toprule
\textbf{Metric} & \textbf{Cataract} & \textbf{DR} & \textbf{Glaucoma} & \textbf{IIR} & \textbf{MAC} & \textbf{MYO} & \textbf{OND} & \textbf{RVD} & \textbf{ROP} \\
\midrule
\textbf{ACC} & 0.54 & 0.71 & 0.49 & 0.19 & 0.56 & 0.66 & 0.83 & 0.21 & 0.65 \\
\bottomrule
\addlinespace[3pt]
\multicolumn{10}{l}{
    \begin{minipage}{0.98\textwidth}
        \textsuperscript{*} \textbf{DR}: Diabetic Retinopathy;
        \textbf{IIR}: Inflammatory/Infectious Retinal Diseases;
        \textbf{MAC}: Macular Diseases;
        \textbf{MYO}: Myopia-Related Diseases;
        \textbf{OND}: Optic Nerve Disorders;
        \textbf{RVD}: Retinal Vascular Diseases;
        \textbf{ROP}: Retinopathy of Prematurity.
    \end{minipage}
}
\end{tabular}
\vspace{-10pt}
\end{table}

\begin{table}[!htbp]
\vspace{-10pt}
\centering
\caption{Performance under different coarse- and fine-grained loss weights $(\lambda_1, \lambda_2)$.}
\vspace{-4pt}
\label{tab:lambda_sweep}
\scriptsize
\setlength{\tabcolsep}{4pt}
\begin{tabular}{cccccc}
\toprule
$\lambda_1$ & $\lambda_2$ & \textbf{LogicEye$^{Score}$} & \textbf{ACC} & \textbf{METEOR} & \textbf{ROUGE-L} \\
\midrule
0 & 0 & 2.31 & 0.41 & 0.46 & 0.43 \\
1 & 0 & 2.68 & 0.49 & \textbf{0.48} & \textbf{0.45} \\
0 & 1 & 2.61 & 0.47 & 0.47 & 0.44 \\
1 & 1 & \textbf{2.92} & \textbf{0.54} & 0.47 & 0.44 \\
1 & 2 & 2.89 & 0.53 & 0.45 & 0.42 \\
2 & 2 & 2.83 & 0.51 & 0.43 & 0.40 \\
\bottomrule
\end{tabular}
\vspace{-15pt}
\end{table}

\noindent \textbf{Sensitivity Analysis of $\lambda_1$ and $\lambda_2$.}
Either classification loss improved LogicEye$^{Score}$ and ACC over generation-only training (Table~\ref{tab:lambda_sweep}). Combining both losses with $\lambda_1=\lambda_2=1$ increased LogicEye$^{Score}$ from 2.31 to 2.92 and ACC from 0.41 to 0.54. This setting yielded the highest LogicEye$^{Score}$ and ACC among the tested weights, with METEOR of 0.47 and ROUGE-L of 0.44.

\vspace{-8pt}
\section{Conclusion}
\vspace{-8pt}
LogicEye$^{Data}$ pairs 804,461 fundus and OCT images with structured pathological reports. LogicEye$^{Agent}$ constructs these reports by checking diagnostic findings and the absence of relevant signs against each image. LogicEye$^{Bench}$ jointly learns report generation and hierarchical disease classification, achieving higher LogicEye$^{Score}$ and diagnostic accuracy than the three evaluated commercial MLLMs on the LogicEye$^{Data}$ test set. The dataset provides fine-grained supervision for training ophthalmic vision-language models to recognize diseases and describe their supporting visual findings.

\bibliographystyle{splncs04} 
\bibliography{ref}

\end{document}